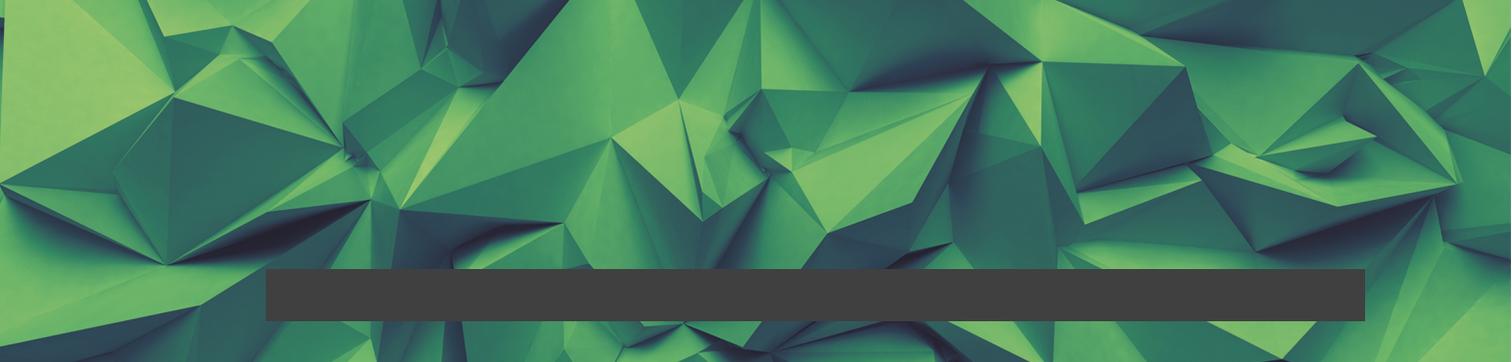

# Interpretable Machine Learning for Privacy-Preserving Pervasive Systems

**Benjamin Baron**
University College London

**Mirco Musolesi**
University College London
and The Alan Turing Institute

Our everyday interactions with pervasive systems generate traces that capture various aspects of human behavior and enable machine learning algorithms to extract latent information about users. In this paper, we propose a machine learning interpretability framework that enables users to understand how these generated traces violate their privacy.

With the emergence of connected devices (*e.g.*, smartphones and smartmeters), pervasive systems generate growing amounts of digital traces as users undergo their everyday activities. These traces are crucial to service providers to understand their customers, to increase the degree of personalization, and enhance the quality of their services. For instance, personal digital traces stemming from public transit smartcards help transportation providers understand the commuting patterns of users; the usage statistics of home appliances can be used to improve energy efficiency; on-street cameras provide police officers with new ways of investigating crimes; content generated through mobile and wearables (e.g., posts in online social media or GPS running routes in specialized websites such as those for fitness) can be used to provide tailored content to individuals; bank transaction logs can be used to spot unusual activity in accounts.

However, sharing these digital traces generated by pervasive systems with service providers might raise concerns with regards to user privacy, as the processing and analysis of these traces can surface latent information about user behaviors. Using machine learning techniques, third parties such as advertisers can identify a single individual from inadequately aggregated datasets shared by service providers either publicly or privately. The common use of ad libraries integrated directly in applications and websites further allows advertisers to collect the same raw traces as the service providers, and infer personal information about users, which can infringe on the users' privacy. In the case of location tracking libraries, these traces might reveal information about the significant places routinely visited by users, which allows to infer a wide range of personal information, including the user's place of residence and work and their future locations.

The main focus of the existing work has been on the performance and interpretability of the techniques to infer personal information and identify users from their digital traces (e.g., Kosinski *et al*. [1]). In particular, there has been a large interest to improve the intelligibility of machine learning models to various audiences, mainly by giving effective and intelligible explanations of the inference task and model to the user [2] [3]. As a result, the explanations must provide an intelligible representation to the users about *what* the model knows and *how* it knows it. With the rise of adversarial and linking attacks on machine learning models, these explanations are important to guarantee the fairness and accountability of the models to the users [4]. Some works have studied the privacy impact of the specific models and proposed methods to improve their interpretability in terms of privacy by allowing the user to adapt the learning and inference algorithms according to their own privacy preferences [5]. However, there have been limited work in relation to how these inference techniques may infringe the user privacy with the personal information they expose. In addition to legal requirements [6], the need for the interpretability through effective explanation of the learning and inference process leading to certain predictions is twofold: (*i*) it helps users understand why their privacy has been violated, and (*ii*) it enables users to trust the model's predictions and recommendations to take the necessary actions to protect their privacy in the future.

In this paper, we discuss the challenges related to the design of an interpretability framework with the goal of supporting interpretation of machine learning techniques that are adopted to infringe the privacy of individuals through personal data inference and user identification. Our contributions are threefold. We state the interpretability and privacy requirements of an effective interpretability framework for privacy-preserving pervasive systems before detailing the functionalities of its components, with a focus on feature selection methods as they are crucial when it comes to present the explanations to the users. We present a case study where we detail a prototype framework that relies on machine learning classifiers with the goal of identifying users from samples of their personal digital traces. Finally, we present the open challenges in this area, discussing a potential research agenda for collaboration across the pervasive computing, human-computer interaction, and machine learning communities.

# TOWARDS A PRIVACY-ORIENTED INTEPRETABILITY FRAMEWORK FOR PERVASIVE SYSTEMS

## Privacy Requirements for Digital Trace Inference Interpretability

The main goal of the proposed interpretability privacy framework is to automatically generate explanations of inference tasks on personal data such that end-users have a better understanding of the privacy risks and implications of these tasks. The explanations should be helpful to the end-users so that they have an understanding on how the underlying machine learning model works and behaves depending on the inputs (*i.e.*, the data provided by the end-user). User understanding is fundamental, as it improves the trust and acceptance of predictions and recommendations given by the model. Further, having a complete mental model of the inference task has been shown to be effective to enable users to better understand the different privacy and security threats [7]. In this section, we present the different requirements concerning the design of a privacy-oriented interpretability framework for machine learning techniques applied to digital traces analysis and inference. Several user studies have been conducted in order to determine the main requirements of explanations about machine learning models and their predictions given to users [2] [3] [5] [7].

The requirements are usually studied from a usability perspective which revolves around user-centric approaches [8]. In particular, researchers and practitioners focus on the so-called *gulf of* *evaluation*, *i.e.*, the gap between representations that can be directly perceived and interpreted by a user provided by a system and her/his expectations and intentions. Another key aspect considered in usability studies is the *gulf of execution*, *i.e.*, the gap between a user's goal related to a specific action and the means to execute that goal. While the main requirements focus on addressing both the gulf of evaluation and the gulf of execution, we argue the case for an additional privacy requirement necessary to provide private explanations. We support the choice of the requirements by providing examples through a discussion of a case study we will present later in the paper. The scenario of the case study describes a machine learning interpretability privacy framework that provides interpretations for inference of personal information from location traces.

**Model understanding (Gulf of evaluation).** There is a general agreement in the community that the explanations given to users should help them have an understanding of the inference task and model, in particular to raise the awareness of the different privacy threats of the inference task. More specifically, the explanations should provide two levels of understanding to users:

- **Why:** The explanations should help users understand individual decisions and predictions given by the model with respect to specific inputs. As so, *instance-based explanations* justify the output of the model, that is providing the reason *why* a specific decision or prediction was given to the end-user. This justification should address the gap between the users' intentions and their perceived functionality of the system. In the case of our interpretability privacy framework for personal information inference, the framework should provide the users with an explanation justifying why the personal information was inferred.
- **How:** The explanation should give the users a comprehensive understanding of the overall machine learning model when possible. Indeed, explaining deep learning algorithms such as Convolutional Neural Networks is still an open problem. *Global explanations* should provide users with a conceptual representation of the model and provide an intuition of *how* it works. Therefore, an essential requirement consists in detailing the main components at the basis of the learning process, in order to help the user understand and identify the privacy threats of the model.

**Model interactions (Gulf of execution).** Once the user has a general understanding of how the model works, they should be able to control and interact with it. In particular, we believe that the explanations should address two following interaction forms, also known as *counterfactual explanations* [3]. These explanations can be subject to probing attacks that aim to perturb the model features.

- **What If:** Understanding how the model behaves depending on different inputs or conditions enables the users to learn how the overall system works. In the case of the interpretability privacy framework for personal information inference, users should be able to determine which information can be inferred depending on the available location information.
- **How To:** Providing explanations and recommendations to the users allows them to have an understanding of which inputs and conditions to change in order to achieve expected predictions or decisions. In the case of the interpretability privacy framework, users may want to know how to prevent any personal information from being disclosed, which would require, for example, the framework to give recommendations on the places and times they should avoid visiting because they may reveal information that was not supposed to be disclosed.

**Privacy-preserving explanations.** When presented to the end-users, the explanations must have specific privacy requirements, in particular, they must preserve the privacy of other users. Since explanations rely on a collection of individual data points or features, they can expose personal information of other users whose digital traces are used to train the inference models. This requirement poses a challenge, as both of the two above requirements may involve exposing a subset of traces from other users to the person who requested the explanation. A possible solution consists in privacy-preserving techniques that can be used to obfuscate or aggregate the information about other users. In the case of the interpretability framework, the explanation that details why personal information was inferred or how the user can be uniquely identified should exclude information about other individual users. In particular, the explanation should not present a user's digital trace that can lead back to the single user who generated it.

## Overview of the Interpretability Privacy Framework

We present an overview of the interpretability privacy framework that addresses the three main requirements presented in the previous section. The framework relies on a distributed architecture, which is depicted in Figure 1. The architecture consists of: (*i*) a privacy-preserving backend and (*ii*) a client that relies on a privacy-preserving library and a series of application plug-ins installed on the end-user's device (*e.g.*, their smartphones) acts as a proxy between the device and the service providers. The distributed architecture implements different components, detailed in the following, that enable users to make informed decisions about sharing their traces that could contain latent personal information.

**Inference component.** The inference component is in charge of inferring latent personal information about a user from their digital traces. This component is implemented in the privacy-preserving backend in order to optimize its performance. The inference component leverages a machine learning model trained with the set of digital traces generated by a large number of users stored in the anonymized data storage. The extent of digital traces generated by users is broad and includes pervasive system traces (*e.g.*, smart meters), location traces (*e.g.*, public transit smartcards or check-ins made on location-based social networks), and on-line activity. The machine learning model of the inference component is specific to the inference task at hand and consists of a classification model that infers personal information such as activity and personal traits from the traces alone. Additionally, the component can further identify the most likely user who could have generated a set of points given as input. In particular, given a set of inputs, the inference task outputs predictions, which consist in a set of activities and personal traits associated with their respective likelihood of belonging to that activity or trait, as well as associated metadata, that is the set of per-user intermediate computation steps involved in the execution of the inference task. The inferred pattern or trait will be the one with the maximum likelihood.

**Explanation component.** The explanation component provides an interface for supporting user interactions and is implemented on the client (for example as an independent application or through a library used by application developers). End-users send queries of pre-defined types to this component, which in turn, translates them into tasks carried out by the remote inference component. The explanation component then retrieves the prediction outputs and metadata of the inference task and provides information about the privacy risk in an intelligible manner. The explanation component should select relevant input features that consist of the user personal digital traces. As we will detail in the following section, the selected features depend on the classifier implemented by the inference component, as well as the domain expertise of the end-user. In order to address the model understanding requirement, the explanation component should also detail the intermediate computation steps involved in the execution of the inference task, given in the prediction metadata. Furthermore, as discussed in the previous section, the explanation component must allow interactions with the end-users. As per the privacy requirement, end-users interested in getting information about their current privacy state and recommendations on possible actions that could affect their privacy state only access to the digital traces they have generated.

**Privacy-preserving component.** This component leverages state-of-the-art privacy-preserving techniques such as differential privacy to share anonymized user traces from the user's device to the remote data store. The goal of this component is to achieve privacy-preserving explanations as detailed above. In particular, with differential privacy, this component can share the user traces and introduce a minimum amount of probabilistic noise in the traces to guarantee a good trade-off between privacy and utility when it comes to use the traces with the inference component [9]. The privacy budget controls the level of privacy provided by the privacy-preserving component. It depends on the service provided, the type of personal information being inferred and the privacy preferences of the user. This results in a user-controllable privacy system, where the user is in charge of finding their optimal trade-off between the level of privacy and the utility of the service. In the case of location data, the component can introduce a certain amount of noise such that users only provide their approximate location. As so, the raw user traces will remain on the user's device until the user makes a decision about sharing their traces or not. This component is necessary in our architecture to guarantee the full anonymity of the traces collected from all the users to perform the inference task.

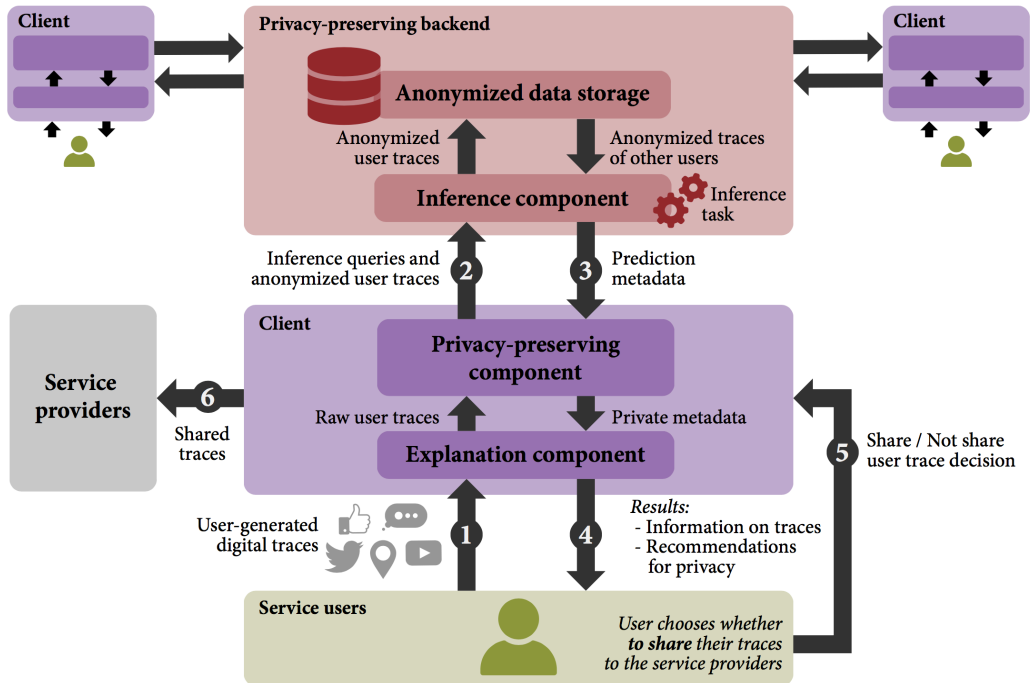

**Figure 1.** Interpretable Machine Learning for Privacy-Preserving Pervasive Systems. **(1)** Our client is installed on the user's device. The privacy-preserving component of the client then analyzes the raw, user-generated content (*e.g.*, Facebook post). **(2)** In turn, the privacy-preserving component client sends the trace in an anonymized way to the remote data storage. **(3)** The inference component then predicts some personal information from the anonymized traces, using the user's traces as well as other users'. **(4)** The prediction is returned to the explanation component, in charge of presenting the inferred personal information to the user. **(5)** The explanation allows for an informed decision about sharing personal information by allowing the user to share or not to share the trace with the service provider. **(6)** If the user chooses to share the trace knowing the privacy implications, the client will share the raw trace with the service provider.

## Implementation of the explanation component with privacy risk and model-dependent feature selection

The goal of the explanation component is to present to the end-user the privacy risk of sharing personal digital traces. The privacy risk corresponds the level of confidence an inference task is able to determine a given personal information from a user trace. To this end, the explanation component computes a score corresponding to the privacy risk based on the input user-generated features and inference metadata. While the presentation of this information to the users is an open research problem [3], possible solutions include numerical scales and the use of visual elements. However, presenting all the features and the metadata would be overwhelming for the user and could hinder the understanding of the model [10]. As such, we argue that the explanation component must select the most relevant features, which, together, contribute the most to the particular inference task requested by the user. We believe that feature selection is the most suitable way for providing an explanation of the inference task [11]. The feature selection method depends on the underlying machine learning model used for the inference component and can be divided into three broad categories [12]. Each method gives a score to the features such that low-scoring features have a low relevance with respect to the other features and should be omitted when presenting an explanation to the end-users. In Table 1 we summarize the different feature selection methods and the different classification models implemented in various the inference components of the literature. Note that our approach is model-dependent, which contrasts with model-agnostic approaches which are limited to instance-based explanations, allowing end-users to only understand why a certain prediction was made. Contrary to this latter approach, our model-dependent approach allows us to provide global explanations of how the whole model works.

Table 1. Feature selection methods according to the inference component used in the literature.

| Inference component | Explanation method for inference | Potential applications |
|---|---|---|
| Cluster matching | Feature rank through entropy measure. | Home activity inference from meter traces [13]. Demographic inference from digital personal traces [14]. |
| Linear regression | Feature rank through coefficients. | Trait and interest inference [1]. |
| Logistic regression | Feature ranking through the results of maximum likelihood estimation. | Trait and interest inference [1]. |
| Naïve Bayes | Feature rank through Information Gain, ReliefF weights, or likelihoods. | Behavior-based identification [15]. Modality-based identification [16]. |
| Multinomial Naïve Bayes | Feature rank through likelihoods and accuracy gain. | Location-based identification [17]. |
| Nearest Neighbors | Feature rank through nearest neighbor distances. | Indoor location inference [18]. |
| Linear Support Vector Machine | Feature rank through support vector weights. | Behavior-based identification [15]. |
| Decision Trees | Feature rank through Information Gain or tree level and frequency. | Behavior-based identification [15]. |
| Random Forest | Feature rank through Gini importance or Information Gain. | Behavior-based identification [15]. |

## CASE STUDY

In this section, we present a case study as an application of our interpretability privacy framework to the problem of personal information prediction from check-in traces such as those used by apps like Swarm/Foursquare using a Multinomial Naïve Bayes classifier [17]. This use case warns users about the privacy implications of a potential check-in, *i.e.*, about the *predictability* of given personal information from a sequence of their location.

Without loss of generality, the inference task will give some personal information about a user such as their political opinion, gender, or religious beliefs. Each personal information has an arbitrary number of classes $F = \{f_1, ..., f_m\}$, for instance in the case of political views, there are two high-level classes $f_1$ and $f_2$ that correspond to liberal and conservative, respectively.

The Multinomial Naïve Bayes classifier computes the probability that a given input (*i.e.*, a set of check-ins given by the end-user $\{c_1, ..., c_n\}$) has of being associated to a certain output (*i.e.*, a personal information class $f_i$). The output with the highest probability becomes the predicted label for the input.

We use $u$ to indicate a single user in the collection of all active Foursquare users $U$, $c_i$ to represent a single check-in as part of the set of check-ins $\boldsymbol{C} = \{c_1, ..., c_n\}$ given by a user, and $f$ to represent a personal information class associated to a user among $\boldsymbol{F} = \{f_1, ..., f_m\}$. A check-in is represented by the tuple (user identifier $u_i$, user personal information $f_i$, location identifier $l_i$). A training set containing the check-ins associated with the personal information of the Foursquare users is fed to the classifier. The inference task consists in determining the correct user personal information $f^\star$ such that it maximizes the following product of the likelihood that each check-in belongs to a user personal information class:

$$f^\star = \operatorname{argmax}_{f \in \boldsymbol{F}} P(f) \prod_{i=1}^{n} P(c_i \mid f).$$

We consider the case of potential personal information prediction from a set of $n$ unlabeled check-ins $\boldsymbol{C}^* = \{c_1^*, ..., c_n^*\}$ using the interpretability privacy framework we described in the previous sections. The check-in sequence is sent to the explanation component, which queries the inference component to determine the most likely personal information class of the Foursquare user who could have generated the check-ins. More specifically, the inference component transmits the individual check-in likelihoods $P(c_i^* \mid f)$ for each check-in $c_i^*$ of $\boldsymbol{C}^*$ and each user personal information class $f$ in the training set. Using this data, the explanation component then presents the explanation of the result of the inference task to the end-user.

We represent an example explanation presented by the framework to the end-user in Figure 2 with labeled Foursquare check-ins collected by Yang *et al.* [19]. Figure 2 shows a possible visualization of the privacy risk using both numerical and chromatic scales. As discussed above, the presentation of this information is an open research challenge, in particular with respect to the explanation of the underlying personal information prediction algorithm. The figure includes statistical information about the user personal information in terms of likelihood, which can be useful to domain expert users, but might be confusing for lay, non-expert users. The explanation further provides a clear recommendation as to whether the user should check-in or not at a given location by estimating the relevance of the new check-in with respect to their current personal information predictability. In the figure, it is apparent that check-in $c_2$ incurs a higher risk than the other check-ins to inferring personal information $F_2$. As we discussed in the previous section, the relevance of an input feature, here a check-in, can be estimated through feature selection methods, in this case, by ranking them according to their likelihoods.

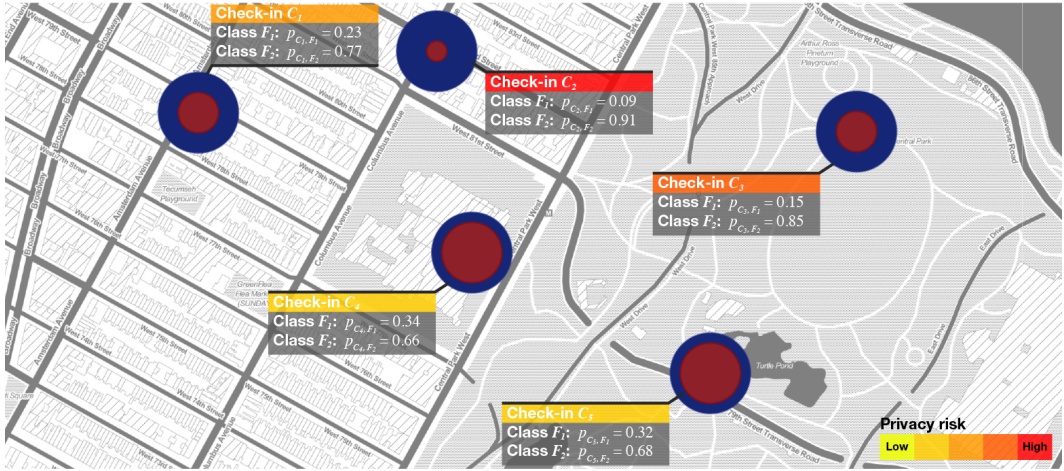

*High privacy risk*

We were able to infer the personal information $F_2$ from the set of check-ins $\mathbf{C}^* = \{C_1, C_2, C_3, C_4, C_5\}$.

**Per-factor likelihood.** The most discriminating check-in locations for $F_2$ are the following (ranked in descending order):
- Check-in $C_2$ because $p_{C_2, F_2} = 0.91$
- Check-in $C_3$ because $p_{C_3, F_2} = 0.85$
- Check-in $C_1$ because $p_{C_1, F_2} = 0.77$
- Check-in $C_5$ because $p_{C_5, F_2} = 0.68$
- Check-in $C_4$ because $p_{C_4, F_2} = 0.66$

**Likelihood product.** The personal information $F_2$ was inferred from the set of check-ins given a prior of 0.68 through the following likelihood products:
- Likelihood product for the personal information $F_1$: $1.222 \times 10^{-5}$
- Likelihood product for the personal information $F_2$: $0.024$

**Figure 2.** Example of an explanation of a personal information prediction task presented to a Foursquare user from the five unlabeled check-ins $\mathbf{C}^* = \{C_1, C_2, C_3, C_4, C_5\}$ among personal information classes $\mathbf{F} = \{F_1, F_2\}$ represented in the figure. In this case, the task explains the personal information prediction $F_1$ to the end-user by presenting the check-ins with a higher privacy risk.

## OPEN RESEARCH CHALLENGES AND OUTLOOK

In this section, we detail the research challenges related to the interpretation and privacy implications of personal information inference tasks from traces generated from the interaction with pervasive systems.

**Information selection.** By default, the model is seen as a black box with a set of inputs and outputs. Predictions alone and metrics based on them do not suffice to characterize and explain the model [2]. While this information is limited, knowing the internals of the model would allow system designers to present relevant information that was learned by the model during its training. The challenge is to determine the amount of information an explanation component should present to users so that they have an understanding of how the model works [2] and what the privacy risks of sharing their data are. This includes determining which features to select and present to the users in order to avoid to give an overwhelming quantity of information, which could result in very complex set of explanations with limited intelligibility. Indeed, while extreme simplicity is not acceptable for users, giving too much information to the user can be overwhelming and will decrease the *quality* of the explanation. For complex models, such as deep learning models, it may be difficult to understand the features and provide a clear explanation to the end-user. In this case, counterfactual explanations or model-agnostic approaches can provide a better solution [3]. With large resulting feature sets, the challenge is to find semantically meaningful feature spaces for given inference tasks. In this case, the use of PCA or SVD could effectively provide lower-dimensional feature sets to be presented to the user.

**Level of expertise of the end-users.** The explanation presented to the users must consider the level of domain expertise of the user, and in particular, the quantity and complexity of information contained in the explanation. As such, an expert user requires extensive explanations, whereas a non-expert user will need simple explanations with a limited amount of information.

The challenge consists in determining the appropriate the level of detail to present to the user. A possible experimental strategy is to conduct user studies in order to allow researchers and practitioners to determine the necessary and sufficient amount of information [2], [5]. The definition of the methodology of these studies is another open challenge *per se* the interpretable machine learning community as discussed below. The challenge is that user studies might be very specific to the particular inference techniques and it might be difficult to extract general conclusions from them [2].

**Definition of privacy-preserving interpretation and anonymity.** When presenting an explanation, sensitive information about other users may be disclosed by the information given in the explanation. The explanation given to a user cannot compromise or expose information about other individuals whose data is contained in the training set of the machine learning model. The explanations can rely on obfuscation techniques such as differential privacy [9] to hide information about other users when presenting an explanation. Alternatively, aggregation methods might guarantee a certain level of anonymity with respect to the trace of other users. However, presenting privacy recommendations to users to help them better protect their privacy requires explanations that contain information about other individuals. This task is particularly challenging, as it requires presenting the general patterns associated to an individual, without revealing the specific patterns associated to other users.

**Recommendations and data shift.** Giving explanations to the users allows them to understand why their privacy can be violated. With recommendations, users can take actions on their future behavior to further protect their privacy. This task is challenging for the following reasons. First, the user must trust the system to follow the recommendation. Effective explanations of the inference task helps increasing user trust in the recommendations. Second, the recommendation given to a specific user depends on the expected future behavior of the other users, as well as whether they follow the recommendations that they received to protect their privacy. As a result, giving recommendations to users may change their behavior and routines, which may in turn invalidate the current training set. This problem is referred to as the data shift problem [20]. A solution consists in training the model again using recently generated traces, but this might not be always possible in practical situations.

**Evaluation of the explanations.** When giving automatically generated explanations to users, we need a methodology to evaluate the explanation interpretability in terms of effectiveness and fidelity of the inference model with respect to the quality of the model understanding and interaction as well as the level of privacy provided (see the privacy requirements we stated in the previous section). Since interpretability depends on a human judgment, it is subjective and depends on the background and level of expertise of the evaluator. Qualitative evaluation is then necessary and can be performed through user studies or surveys. In this case, the challenge is to design a methodology to test and assess the subjective comprehensibility of the explanation with user studies by choosing the right set of users, including expert users and non-expert users. One way of carrying out the evaluation would be to present different explanation framing strategies to different end-users, using for example A/B testing. In particular, it is important to determine the performance metrics used to reliably evaluate the explanation. For instance, trust in the explanation is a common metric that is highly subjective. While qualitative evaluations are necessary, an interesting research question is how to perform systematic quantitative evaluation of explanations given target performance metrics, without the involvement of users [10].

We believe that addressing these research challenges will be possible only through the collaboration of researchers from different communities, including, but not limited to, pervasive computing, human-computer interaction, data visualization and machine learning.

## ABOUT THE AUTHORS

**Benjamin Baron** is a Postdoctoral Research Fellow at University College London. His interests include digital traces analytics, mobile phone data modeling and security&privacy. Dr Baron has a Ph.D. in Computer Science from Sorbonne Université.

**Mirco Musolesi** is a Reader in Data Science at University College London and Turing Fellow at The Alan Turing Institute. His research interests include ubiquitous computing, intelligent systems and security&privacy. Dr Musolesi has a Ph.D. in Computer Science from University College London.